\pdfoutput=1
\documentclass{article}

\usepackage[usenames,dvipsnames]{xcolor}
\usepackage{microtype}
\usepackage{graphicx}
\usepackage{subfigure}
\usepackage{booktabs} % for professional tables

\usepackage{hyperref}

\usepackage[utf8]{inputenc} % allow utf-8 input
\usepackage[T1]{fontenc}    % use 8-bit T1 fonts
\usepackage{nicefrac}       % compact symbols for 1/2, etc.
\usepackage{times}

\usepackage{amsthm,amsmath,amssymb,amsfonts,epsfig,pgf,rotating,longtable,booktabs, colortbl}
\usepackage{tabularx}
\usepackage{wrapfig}
\usepackage{multicol}
\usepackage{subfigure}
\usepackage{multirow}

\newtheorem{thm}{Proposition}

\usepackage[font=small,labelfont=bf,tableposition=top]{caption}

\usepackage[accepted]{icml2021}

\icmltitlerunning{Variational Rejection Particle Filtering}

\begin{document}

\twocolumn[
\icmltitle{Variational Rejection Particle Filtering}



\icmlsetsymbol{equal}{*}

\begin{icmlauthorlist}
\icmlauthor{Rahul Sharma}{iitc}
\icmlauthor{Soumya Banerjee}{iitc}
\icmlauthor{Dootika Vats}{iitm}
\icmlauthor{Piyush Rai}{iitc}
\end{icmlauthorlist}

\icmlaffiliation{iitc}{Department of Computer Science and Engineering, IIT Kanpur}
\icmlaffiliation{iitm}{ Department of Mathematics and Statistics, IIT Kanpur}

\icmlcorrespondingauthor{Rahul Sharma}{rsharma@cse.iitk.ac.in}
%\icmlcorrespondingauthor{Eee Pppp}{ep@eden.co.uk}

\icmlkeywords{Machine Learning, ICML}

\vskip 0.3in
]



\printAffiliationsAndNotice{}  % leave blank if no need to mention equal contribution

\begin{abstract}
%\textbf{change the abstract, too similar to introduction}
%Effective variational inference (VI) crucially depends on a flexible family of distribution. In the existing literature, there are two main approaches to construct a flexible VI bound: particle-based approximations (sequential Monte-Carlo (SMC), importance sampling (IS)), and sampling-based methods (Rejection Sampling, Markov Chain Monte-Carlo (MCMC)). 
%In this paper, we propose a new family of variational bounds Variational Rejection Particle Filtering (VRPF) that combines particle filters with a sampling-based approach. In particular, VRPF employs an \emph{approximate} rejection sampling step within a particle filter to construct a flexible family of variational inference (VI) distributions. Furthermore, to improve felxibility, we include a resampling step within VRPF via Bernoulli race: a generalization of a  Bernoulli factory. The VRPF lower bound can be optimized efficiently with respect to the variational parameters and generalizes several existing approaches in the VI literature into a single framework. We report theoretical properties of the lower bound and demonstrate experiments on various sequential models, such as the Gaussian state-space model and variational recurrent neural net (VRNN), on which VRPF empirically outperforms the existing state-of-the-art methods.

We present a variational inference (VI) framework that unifies and leverages sequential Monte-Carlo (particle filtering) with \emph{approximate} rejection sampling to construct a flexible family of variational distributions. Furthermore, we augment this approach with a resampling step via Bernoulli race, a generalization of a Bernoulli factory, to obtain a low-variance estimator of the marginal likelihood. Our framework, Variational Rejection Particle Filtering (VRPF), leads to novel variational bounds on the marginal likelihood, which can be optimized efficiently with respect to the variational parameters and generalizes several existing approaches in the VI literature. We also present theoretical properties of the variational bound and demonstrate experiments on various models of sequential data, such as the Gaussian state-space model and variational recurrent neural net (VRNN), on which VRPF outperforms various existing state-of-the-art VI methods.

%Recent work has explored sequential Monte-Carlo (SMC) methods to construct variational distributions, which can, in principle, approximate the target posterior arbitrarily well, which is especially appealing for models with inherent sequential structure. However, SMC, which represents the posterior using a weighted set of particles, often suffers from particle weight degeneracy, leading to a large variance of the resulting estimators. To address this issue, we present a novel approach that leverages the idea of \emph{partial} rejection control (PRC) for developing a robust variational inference (VI) framework. In addition to developing a superior VI bound, we propose a novel marginal likelihood estimator constructed via a Bernoulli race: a generalization of the Bernoulli factory to construct unbiased estimators for SMC-PRC. The resulting variational lower bound can be optimized efficiently with respect to the variational parameters and generalizes several existing approaches in the VI literature into a single framework. We show theoretical properties of the lower bound and report experiments on various sequential models, such as the Gaussian state-space model and variational RNN, on which our approach outperforms existing methods.
\end{abstract}

\section{Introduction}
Exact inference in latent variable models (LVM) is usually intractable. Recently VI~\citep{blei2017variational} methods have received considerable interest for LVMs due to their excellent scalability on large-scale datasets. This is made possible thanks to scalable amortized VI methods \citep{kingma2013auto,ranganath2014black} and stochastic VI~\citep{hoffman2013stochastic}. In particular, VI maximizes a lower bound on the log marginal likelihood to obtain an approximate posterior. Constructing a low variance estimator of the marginal likelihood is desirable due to a tighter VI bound and yields a better approximation to the target posterior~\citep{domke2019divide}. The two basic schemes for constructing a low variance estimator of the marginal likelihood are based on sampling methods (MCMC, rejection sampling)~\cite{salimans2015markov,ruiz2019contrastive,hoffman2017learning,grover2018variational} or particle-based approximation (e.g., sequential Monte-Carlo (SMC) and importance sampling (IS))~\cite{burda2015importance,maddison2017filtering}.

% We focus on leveraging the merits of using a particle approximation in conjunction with a  sampling-based approach in order to obtain tighter VI bounds.

%Therefore, a natural question that arises from the existing works is whether we can formulate a VI bound that combines a particle approximation and a sampling-based approach. 

%\textbf{\textcolor{red}{start with positive then put negative} Instead of relying only on particles or sampling, we would like to combine these two approaches synergistically, developing a flexible family of approximate posteriors.}  Instead of relying only on particles or sampling, we would like to combine these two approaches synergistically, developing a flexible family of approximate posteriors

%However, combining these two approaches is a challenging task as sampling-based methods often induce intractability. In this work, we develop a unified approach that combines particle approximation with a sampling-based approach.

%\textbf{rewrite this para}
In this paper, we develop a novel VI bound, VRPF (Variational Rejection Particle Filtering), which is based on unifying a particle approximation with a  sampling-based approach. In particular, the proposed bound formulates an efficient variational proposal via particle approximation followed by further refinement through a sampling-based technique. To accomplish this, we leverage the idea of \emph{partial} rejection control (PRC)~\citep{peters2012sequential,liu1998rejection}, an \emph{approximate} rejection sampling step, within the framework of SMC, which is a particle approximation, and exploit their synergy to develop a flexible family of approximate posteriors. Note that VRPF does not employ sampling in a traditional sense; instead, it uses a greedy form of sampling called \emph{partial} accept-reject. Given a sequence of samples, VRPF applies accept-reject only on the most recent update, consequently increasing the sampling efficiency for high-dimensional sequences.

%It is well known in the literature that SMC-PRC produces a low variance estimator of the marginal likelihood~\cite{peters2012sequential,kudlicka2020particle}. 

%In contrast to standard sampling algorithms that may reject the \emph{entire} stream of particles, SMC-PRC uses a \emph{partial} accept-reject on the most recent update increasing the sampling efficiency. Further, the variational framework of SMC-PRC is interesting in itself as it combines accept-reject with particle filter methods. Therefore, our proposed bound VRPF generalizes several existing approaches for example: Variational Rejection Sampling (VRS)~\cite{grover2018variational}, FIVO~\cite{maddison2014sampling}, IWAE~\cite{burda2015importance}, and standard variational Bayes~\cite{blei2017variational}. 

%Recall that constructing a low variance estimator of the marginal likelihood is significant due to a better approximation of the posterior density. Therefore, the key technical obstacle in formulating VRPF bound is to perform unbiased resampling.

%\textbf{doubtful: Providing the complete solution,just define the problem}
Constructing a VRPF bound is non-trivial because of the computational intractability induced by the sampling-based methods within particle approximation. In particular, the use of \emph{partial} accept-reject makes the particle weights intractable. Since the weights are not analytically available, we cannot exploit variance reduction properties of resampling~\citep{doucet2009tutorial}. To alleviate this issue, we further employ a Bernoulli race algorithm~\cite{dughmi2017bernoulli,schmon2019bernoulli} to perform unbiased resampling. Specifically, for VRPF, we first construct an unbiased estimator of the particle weights, followed by resampling via Bernoulli race. Note that the VRPF bound takes advantage of resampling in addition to accept-reject, therefore formulating a flexible family of VI bounds.

\textbf{Summary of Contributions.} The main contributions of this paper are summarized as follows

\vspace{-4mm}
\begin{itemize}
    \item We construct a novel VI bound, VRPF, which unifies a sampling-based method with a particle approximation in a theoretically consistent manner. Specifically, we formulate a VI bound that leverages the benefits of SMC, which is a particle approximation, and PRC, which a sampling-based approach. We perform experiments on a Gaussian state-space model (SSM) and variational recurrent neural networks (VRNN) on which our method outperforms existing state-of-the-art methods like FIVO~\citep{maddison2017filtering} VSMC~\citep{naesseth2017variational} and IWAE~\citep{burda2015importance}.
    \vspace{-2mm}
    \item Another key aspect of our approach is the use of \emph{partial}-sampling. Through detailed experiments, we show that \emph{partial}-sampling is useful despite using a simple accept-reject technique. In particular, we would like to highlight that the sequence length in our experiments is fairly high-dimensional, i.e. of the order $ 10^{6} $. Therefore, our work provides interesting insights and motivation on the exploration of \emph{partial}/greedy-sampling for high-dimensional time series models. 
    \vspace{-2mm}
    \item We add to the existing line of work on unbiased estimation of the marginal likelihood for general SMC with PRC, building on the works of~\citet{kudlicka2020particle}. We provide an unbiased estimator of the marginal likelihood of~\citet{peters2012sequential} by demonstrating that \citet{peters2012sequential} is a special case of Bernoulli Race Particle Filter (BRPF)~\cite{schmon2019bernoulli}.  
    % Apart from VI, the proposed estimator of the marginal likelihood might be useful for particle-MCMC~\citep{andrieu2010particle} due to its low variance. 
    % However, we would like to highlight that~\citet{kudlicka2020particle} doesn't consider the exact construction of~\citet{peters2012sequential}.
\end{itemize}
 
 \textbf{Outline.} The rest of the paper is organized as follows: In Section~\ref{sec_background}, we provide a brief review on SMC with Partial Rejection Control (SMC-PRC) and Bernoulli race. Section~\ref{section_VRPF} introduces VRPF bound and presents theoretical results about the Monte-Carlo estimator and efficient ways to optimize it. Finally, we discuss related work and present experiments on the Gaussian SSM and VRNN.

\section{Background}\label{sec_background}

%\textbf{include some explanation on SSM as done in AE-SMC paper}
Consider a state-space model (SSM) over a set of latent variables $ z_{1:T}=(z_{1},z_{2},\hdots,z_{T})$ and real-valued observations $x_{1:T}=(x_{1},x_{2},\hdots,x_{T})$. We are interested in inferring the posterior distribution of the latent variables, i.e., $ p_{\theta}(z_{1:T}|x_{1:T})$ where $\theta$ represents the model parameters. The task is, in general, intractable. In such sequential models, SMC is a valuable tool for approximating the posterior distribution~\citep{doucet2009tutorial}.

\textbf{Notation}: For the rest of the paper we use some common notations from SMC and VI literature where $ z^{i}_{t}$ denotes the $ i^{th} $ particle at time $ t $, $ A^{i}_{t-1} $ is the ancestor variable for the $ i^{th} $ particle at time $ t $, and  $ \theta$ and $\phi$ are model and variational parameters, respectively.

%We further assume that the joint density $ p_{\theta}(x_{1:t},z_{1:t}) $ factorizes as follows for $ t \in \{2,3,\hdots,T\}$
%\begin{equation}
%    p_{\theta}(x_{1:t},z_{1:t}) = p_{\theta}(x_{1:t-1},z_{1:t-1})  p_{\theta}(x_{t},z_{t}|x_{1:t-1},z_{1:t-1}). \label{joint_dist}
%\end{equation}

\subsection{Sequential Monte Carlo with Partial Rejection Control (SMC-PRC)}
\label{sec:smc} 
An SMC sampler approximates a sequence of densities $ \left\{p_{\theta}(z_{1:t}|x_{1:t})\right\}_{t=1}^{T}$ through a set of $N$ weighted samples  generated from a proposal distribution. Let  the proposal density be
\vspace{-1mm}
\begin{equation}
    q_{\phi}(z_{1:T}|x_{1:T}) = \prod_{t=1}^{T}q_{\phi}(z_{t}|x_{1:t},z_{1:t-1}). \label{var_post}
\end{equation}
Consider time $t-1$ at which we have uniformly weighted samples $ \{N^{-1},z^{i}_{1:t-1},A^{i}_{t-1}\}_{i=1}^{N}$ estimating $p_{\theta}(z_{1:t-1}|x_{1:t-1}) $. We want to estimate  $ p_{\theta}(z_{1:t}|x_{1:t}) $ such that particles with low importance weights are automatically rejected. PRC achieves this by using an \emph{approximate} rejection sampling step. The overall procedure is as follows:
%Note that we have employed a continuous acceptance probability for PRC (see equation~\eqref{accept_prob}) closely related to Barker's acceptance probability~\cite{minh2015understanding} also used in~\cite{grover2018variational}.
%
\begin{enumerate}
    \item For $i=1,2,\hdots,N$, generate 
    \[ z^{i}_{t}\sim q_{\phi} \left(z_{t}|x_{1:t},z^{A^{i}_{t-1}}_{1:t-1} \right).
    \]
    \vspace{-2mm}
    \item Accept $ z^{i}_{t} $ with probability $ a_{\theta,\phi}\big(z^{i}_{t}|z^{A^{i}_{t-1}}_{1:t-1},x_{1:t} \big) =$
    \vspace{-2mm}
    \begin{equation}
        \left(1+\frac{M(i,t-1)q_{\phi} \left(z^{i}_{t}|x_{1:t},z^{A^{i}_{t-1}}_{1:t-1} \right)}{p_{\theta} \left(x_{t},z^{i}_{t}|x_{1:t-1},z^{A^{i}_{t-1}}_{1:t-1} \right)} \right)^{-1}, \label{accept_prob}
    \end{equation}
    where $ M(i,t-1) $ is a hyperparameter controlling the acceptance rate (see Proposition~\ref{prop_3} and Section~\ref{tune_M} for more details). Note that PRC applies accept-reject only on $z^{i}_{t}$, not on the entire trajectory.
    \item If $ z^{i}_{t} $ is rejected go to step 1.
   % \item Using the factorization assumption of Eq.~\eqref{joint_dist}, approximate $ p_{\theta}(z_{1:t}|x_{1:t}) $ as follows
%    \begin{equation}
%        p_{\theta}(z_{1:t}|x_{1:t}) \approx \sum_{i=1}^{N}w^{i}_{t} \delta_{z_{1:t}^{i}}, \label{eq_recur}
%    \end{equation}
%    where $\delta$ is the Dirac-delta function, and the new weights $w^{i}_{t} $ are given by
%    \begin{equation}
%        w^{i}_{t}\propto w^{i}_{t-1}\frac{p_{\theta}(x_{t},z^{i}_{t}|x_{1:t-1},z^{i}_{1:t-1})}{q_{\phi}(z^{i}_{t}|x_{1:t},z^{i}_{1:t-1})} \label{new_weights}
%    \end{equation}
    \item The new incremental importance weight of the accepted sample is
    \vspace{-2mm}
    \begin{equation}
        \alpha_{t} \left(z^{i}_{1:t}\right)= c^{i}_{t} Z\left(z^{A^{i}_{t-1}}_{1:t-1},x_{1:t} \right) , \label{iiw}
    \end{equation}
    where $ c^{i}_{t} $ is
    \vspace{-2mm}
\begin{equation}
    c^{i}_{t} = \frac{p_{\theta} \left(x_{t},z^{i}_{t}|x_{1:t-1},z^{A^{i}_{t-1}}_{1:t-1} \right) }{q_{\phi} \left(z^{i}_{t}|x_{1:t},z^{A^{i}_{t-1}}_{1:t-1} \right) a_{\theta,\phi} \left(z^{i}_{t}|z^{A^{i}_{t-1}}_{1:t-1},x_{1:t} \right)}, \label{mult_nom_const}
\end{equation}
and the intractable normalization constant $ Z(.) $  (For simplicity of notation, we ignore the dependence of $ Z(.) $ on $ M(i,t-1) $)
\vspace{-2mm}
\begin{equation}
   Z \left(z^{A^{i}_{t-1}}_{1:t-1},x_{1:t} \right) = \mathbb{E} \left[a_{\theta,\phi} \left(z_{t}|z^{A^{i}_{t-1}}_{1:t-1},x_{1:t} \right)\right].
\end{equation}

%{\color{red}(DV): Something is missing here. This sentence doesn't make sense}

    \item Compute the Monte-Carlo estimator of the unnormalized weights %{\color{red}(DV): why does below have notation $\tilde{w}$? This hasn't been introduced before}
    \vspace{-2mm}
\begin{equation}
     \widetilde{w}^{i}_{t} =    \frac{c^{i}_{t}}{K}\sum_{k=1}^{K} a_{\theta,\phi} \left(\delta^{i,k}_{t}|z^{A^{i}_{t-1}}_{1:t-1},x_{1:t} \right)   \label{unnorm_weight},
\end{equation}
where 
\[
\delta^{i,k}_{t}\sim q_{\phi}(z_{t}|x_{1:t},z^{A^{i}_{t-1}}_{1:t-1}). \] 
Note that $ \widetilde{w}^{i}_{t}$ is essential for constructing an unbiased estimator of $ p_{\theta}(x_{1:T}) $.
    \item Using a Bernoulli race algorithm described in Section~\ref{subsec:bernoulli_race}, generate %
    \vspace{-2mm}
  \begin{equation}
    A^{i}_{t} \sim  \text{Multinoulli}\left(\frac{\alpha_{t} (z^{i}_{1:t})}{\sum_{j=1}^{N}\alpha_{t}(z^{j}_{1:t})} \right)_{i=1}^{N}. \label{ancestor_sample} 
  \end{equation}

%  \item Set the new weights $ w^{i}_{t}=\frac{1}{N} $ and $ z^{i}_{t} = z^{A^{i}_{t}}_{t}$ for $i=1,2,\hdots,N$. 
%  \item Using the factorization assumption of Eq.~\eqref{joint_dist}, approximate $ p_{\theta}(z_{1:t}|x_{1:t}) $ as follows
%   \begin{equation}
%        p_{\theta}(z_{1:t}|x_{1:t}) \approx \sum_{i=1}^{N}w^{i}_{t} \delta_{z_{1:t}^{i}}, \label{eq_recur}
%    \end{equation}
%    where $\delta$ is the Dirac-delta function
\end{enumerate}

\begin{algorithm*}[h]
\caption{Estimating the VRPF lower bound}
\label{Algorithm_1}
\begin{multicols}{2}
\begin{algorithmic}[1] 
\STATE{\textbf{Required}: $ N $, $ K $, and $ M $ }
\FOR{$ t \in \{1,2,\hdots,T\} $ }
\FOR{$ i \in \{1,2,\hdots,N \} $}
\STATE{$ z_{t}^{i}, c^{i}_{t},\widetilde{w}^{i}_{t} \sim \textbf{PRC}\left(q,p,M(i,t-1) \right) $}
%\vspace{-3mm}
\STATE {$ z_{1:t}^{i} = \big(z_{1:t-1}^{A^{i}_{t-1}},z_{t}^{i} \big)  $   }
\ENDFOR
%\STATE{ $ \{w_{t}^{i}\}_{i=1}^{N}\propto \{ \frac{1}{N}\alpha_{t}(z_{1:t}^{i}) \}_{i=1}^{N} $ }
\FOR{$ i \in \{1,2,\hdots,N \} $}
\STATE{$ A^{i}_{t} = \text{\textbf{BR}}\left( \{c_{t}^{i},z_{1:t}^{i}\}_{i=1}^{N} \right) $}
\ENDFOR
\ENDFOR
\STATE{\textbf{return} $\log \prod_{t=1}^{T}\left(  \frac{1}{N}\sum_{i=1}^{N}\widetilde{w}^{i}_{t} \right)$ }
\STATE{}
\STATE{\textbf{PRC} $ \left(q,p,M(i,t-1) \right) $}
\WHILE{sample not accepted}
\STATE{Generate $ z_{t}^{i} \sim$ $ q_{\phi} \big(z_{t}|x_{1:t},z_{1:t-1}^{A^{i}_{t-1}} \big) $}
\STATE{Accept $ z_{t}^{i}$ with probability  $a_{\theta,\phi} \big(z^{i}_{t}|z^{A^{i}_{t-1}}_{1:t-1},x_{1:t}\big)$ }
\ENDWHILE
\STATE{ Sample $ \{\delta^{i,k}_{t}\}_{k=1}^{K} \sim q_{\phi} \big(z_{t}|x_{1:t},z_{1:t-1}^{A^{i}_{t-1}} \big) $ }
\STATE{Calculate $ \widetilde{w}^{i}_{t} $ from Eq.~\eqref{unnorm_weight} }
\STATE{ Calculate $c^{i}_{t} $ from Eq.~\eqref{mult_nom_const} }
\STATE{\textbf{return} $\left( z^{i}_{t}, c^{i}_{t}, \widetilde{w}^{i}_{t} \right) $ }
\STATE{}
\STATE{\textbf{BR} $ \left( \{c_{t}^{i},z_{1:t}^{i}\}_{i=1}^{N} \right)$ }
\vspace{-1mm}
\STATE{ Sample $ C \sim \text{Multinoulli}\left(\frac{c^{i}_{t}}{\sum_{j=1}^{N}c^{j}_{t} } \right)_{i=1}^{N}$ }
\vspace{-1mm}
\IF{ $ C == i $ }
\STATE{ Sample $ U_{i} \sim U[0,1] $}
%\vspace{-1mm}
\STATE{ $ \kappa_{t} \sim$ $ q_{\phi} \big(z_{t}|x_{1:t},z_{1:t-1}^{A^{i}_{t-1}} \big) $ }
\ENDIF
\IF{ $ U_{i} < a_{\theta,\phi}(\kappa_{t}|z^{A^{i}_{t-1}}_{1:t-1},x_{1:t}) $ }
\STATE{\textbf{return}  $ ( i ) $}
\ELSE
\STATE{ return \textbf{BR} $ \left( \{c_{t}^{i},z_{1:t}^{i}\}_{i=1}^{N} \right)$ }
\ENDIF
\end{algorithmic}
\end{multicols} 
\vspace{-4mm}
\end{algorithm*}

Simulation of ancestor variables in Eq.~\eqref{ancestor_sample} is non-trivial  due to intractable normalization constants in the incremental importance weight (see \eqref{iiw}). Vanilla Monte-Carlo estimation of $ \alpha_{t}(.) $ yields biased samples of ancestor variables from Eq.~\eqref{ancestor_sample}. To address this issue, we leverage a generalization of Bernoulli factory~\citep{asmussen1992stationarity}, called Bernoulli race.

\subsection{Bernoulli Race}
\label{subsec:bernoulli_race}
Suppose we can simulate Bernoulli$ (p^{i}_{t}) $ outcomes where $ p^{i}_{t} $ are intractable. Bernoulli factories  simulate an event of probability $ f(p^{i}_{t}) $, where $ f(.) $ is some desired function. In our case, the intractable coin probability $ p^{i}_{t} $ is the intractable normalization constant,
\vspace{-1mm}
\begin{equation}
    p^{i}_{t}  = Z \left(z^{A^{i}_{t-1}}_{1:t-1},x_{1:t} \right). \label{Bern_coin}
\end{equation}
Since $ p^{i}_{t} \in [0,1] $ and we can easily simulate this coin, we obtain the Bernoulli race algorithm below.

\begin{enumerate}
\item Required: Constants $ \{c^{i}_{t}\}_{i=1}^{N} $ see \eqref{mult_nom_const}.
    \item Sample $ C \sim \text{Categorical}\left(\frac{c^{1}_{t}}{\sum_{j=1}^{N}c^{j}_{t} },\hdots,\frac{c^{N}_{t}}{\sum_{j=1}^{N}c^{j}_{t} } \right)$
    \item If $ C = i $, independently generate $ U_{i}\sim U[0,1]$ and 
    \vspace{-1mm}
    \[\kappa_{t} \sim  q_{\phi} \left(z_{t}|x_{1:t},z^{A^{i}_{t-1}}_{1:t-1} \right) 
    \]
    \vspace{-3mm}
    \begin{itemize}
        \item If $ U_{i}<a_{\theta,\phi}(\kappa_{t}|z^{A^{i}_{t-1}}_{1:t-1},x_{1:t}) $ output $ i $
        \item Else go to step 2
    \end{itemize}
\end{enumerate}

The Bernoulli race produces unbiased ancestor variables. Further we can easily control the efficiency of the proposed Bernoulli race through the  hyper-parameter $ M $ (as in~\eqref{accept_prob}).

%\textbf{Add some more explanation on how SMC-PRC is a special case}
Note that we have replaced the true intractable weights with their Monte-Carlo estimator and performed resampling through the Bernoulli race. Therefore, it is easy to see that SMC-PRC is indeed a particular case of BRPF. Another interesting aspect about SMC-PRC is that we can easily control its efficiency through hyper-parameter $ M $. For more details regarding efficient implementation, see Section~\ref{tune_M}. 

%\textbf{We can say something about the efficiency of Monte-Carlo estimator and hyper-parameter M}
 % in contrast to existing Bernoulli factory algorithms~\cite{schmon2019bernoulli}. For details on efficiency and correctness, please refer to Section~\eqref{theore_prop} and Section~\eqref{tune_M}.

%\newpage

\section{Variational Rejection Particle Filtering}\label{section_VRPF}
Our proposed VRPF bound is constructed through a marginal likelihood estimator obtained by combining the SMC sampler with a PRC step and Bernoulli race. Note that the variance of estimators obtained through SMC-PRC particle filter is usually low~\cite{peters2012sequential,kudlicka2020particle}. Therefore, we expect VRPF to be a tighter bound in general~\citep{domke2019divide} compared to the standard SMC based bounds used in recent works~\cite{maddison2017filtering,naesseth2017variational,le2017auto}. Algorithm~\ref{Algorithm_1} summarizes the generative process to simulate the VRPF bound and Figure~\ref{fig:VRPF_visual} presents a visualization of the VRPF generative process.
%\textbf{Explain More}

We now demonstrate how to leverage PRC to develop a robust VI framework (VRPF). Specifically, the proposed framework formulates a VI lower bound via a marginal likelihood estimator obtained through Algorithm~\ref{Algorithm_1}. Let the sampling distribution of Algorithm~\ref{Algorithm_1} be $ Q_{\text{VRPF}}$ with variational parameters $ \phi $ and model parameters $ \theta $. If $K$ are the Monte-Carlo samples used for estimating $ Z(.) $, the VRPF bound is
% 
%Our framework is based on the VRPF bound presented below. The Monte-Carlo estimator of VRPF bound is 
% 
%The complete sampling distribution of Algorithm~\eqref{Algorithm_1} is as follows.
%\vspace{-1mm}
%\begin{equation}
%\begin{split}
%     &  Q_{\text{VRPF} } \left(z^{1:N}_{1:T},A^{1:N}_{1:T-1} ,\delta^{1:N,1:K}_{1:T} \right)    =   
% \left(\prod_{k=1}^{K}\prod_{i=1}^{N} q_{\phi}(\delta^{i,k}_{1}|x_{1}) \prod_{t=2}^{T} \prod_{i=1}^{N}\prod_{k=1}^{K} q_{\phi}(\delta^{i,k}_{t}|x_{1:t},z^{A^{i}_{t-1}}_{1:t-1})\right) \times \\
% &    \left( \prod_{i=1}^{N}\frac{q_{\phi}(z^{i}_{1}|x_{1})a_{\theta,\phi}(z^{i}_{1}|x_{1})} {Z(x_{1})} 
%    \prod_{t=1}^{T-1} \prod_{i=1}^{N} \text{Discrete}(A^{i}_{t}|\alpha_{t})   \frac{q_{\phi}(z^{i}_{t+1}|x_{1:t+1},z^{A^{i}_{t}}_{1:t})a_{\theta,\phi}(z^{i}_{t+1}|z^{A^{i}_{t}}_{1:t},x_{1:t+1})} {Z(z^{A^{i}_{t}}_{1:t},x_{1:t+1})} \right)
%    \end{split}
%    \label{eq:sampl_dist}
%\end{equation}
% 
%The normalization constants $ Z(.) $ in Eq.~\eqref{eq:sampl_dist} are intractable and have to be estimated while calculating the weights. Therefore, we introduce an extra parameter $ K $, denoting the number of Monte-Carlo samples used to estimate $ Z(.) $. 
% 
%Let the sampling distribution of Algorithm~\eqref{Algorithm_1} be $ Q_{\text{VRPF}}$ with variational parameters $ \phi $ and model parameters $ \theta $. Assuming that we have used $ K $ Monte-Carlo samples for estimating $ Z(.) $ we can write down the VRPF bound as follows:
% 
% \vspace{-2mm}
\begin{equation}
    \mathcal{L}_{\text{VRPF}}(\theta,\phi;x_{1:T},K)=\mathbb{E}_{Q_{\text{VRPF} }  } \left[  \sum_{t=1}^{T} \log \frac{1}{N} \sum_{i=1}^{N}\widetilde{w}^{i}_{t} \right].  \label{VRPF-bound}
\end{equation}

Note that many hyper-parameters affect the VRPF bound: the number of particles ($ N $), the number of Monte-Carlo samples ($ K $), and the accept-reject constant ($ M $). We have discussed the effect of each hyper-parameter in Section~\ref{theore_prop}.  Section~\ref{grad_sec} discusses the gradient estimation of VRPF bound. Finally, we explain how to tune $ M $ in Section~\ref{tune_M}. Note that tuning $ M $ is crucial as it affects the efficiency of the PRC step as well as the Bernoulli race. Due to intractability, we will construct a Monte-Carlo estimate of  $ \mathcal{L}_{\text{VRPF}} $ by drawing a sample from $ Q_{\text{VRPF}}$.
\vspace{-2mm}
\begin{equation}
    \hat{\mathcal{L}}_{\text{VRPF}}(\theta,\phi;x_{1:T},K)= \sum_{t=1}^{T} \log \frac{1}{N} \sum_{i=1}^{N}\widetilde{w}^{i}_{t}.  \label{VRPF-MC}
\end{equation}
 Using Jensen's inequality and unbiasedness of $\exp(\hat{\mathcal{L}}_{\text{VRPF}}) $ (see Proposition~\ref{prop_2}), we can show that $ \mathbb{E}[\hat{\mathcal{L}}_{\text{VRPF}}] $ is a lower bound on the log marginal likelihood. We maximize the VRPF bound with respect to model parameters $ \theta $ and variational parameters $ \phi $. This requires estimating the gradient  the  details of which are provided in  Section~\ref{grad_sec}.
%We maximize the VRPF bound with respect to model parameters $ \theta $ and variational parameters $ \phi $. This requires estimating the gradient  the  details of which are provided in  Section~\eqref{grad_sec}.
%There are three hyper-parameters that affect the Monte-Carlo estimate $  \hat{\mathcal{L}}_{\text{VRPF}} $ : $N, M $ and $ K $. Role of $ N $ has already been discussed in the literature~\cite{berard2014lognormal,naesseth2017variational}. As $ N $ increases, we expect the VRPF bound to get tighter. Hence, we will focus our attention on  $ M $ and $ K $.

\subsection{Theoretical Properties}\label{theore_prop}
We now present properties of the Monte-Carlo estimator  $ \hat{ \mathcal{L}}_{\text{VRPF}} $. The key variables that affect this estimator are the number of samples, $ N $, hyper-parameter $ M $, and the number of Monte-Carlo samples used to compute the normalization constant $ Z(.) $, i.e., $ K$. As discussed by~\citet{berard2014lognormal,naesseth2017variational}, as $ N $ increases, we expect the VRPF bound to get tighter. Hence, we will focus our attention on  $ M $ and $ K $. All the proofs can be found in the supplementary material.

\begin{thm}
Bernoulli race produces unbiased ancestor variables. Further, let $ \Lambda_{t} $ be the number of iterations required for generating one ancestor variable, then $  \Lambda _{t} \sim \text{Geom}\left( \mathbb{E}[\Lambda_{t}]^{-1} \right) $ where
\begin{equation}\nonumber
\mathbb{E}[\Lambda_{t}] = \frac{\sum_{i=1}^{N}c^{i}_{t} }{ \sum_{i=1}^{N}c^{i}_{t}Z(z^{A^{i}_{t-1}}_{1:t-1},x_{1:t}) }.
\end{equation}

%\begin{itemize}
%    \item The average execution time for Bernoulli race is given by:
%    \begin{equation}
        
%    \end{equation}
%\end{itemize}
\label{prop_1}
\end{thm}
\vspace{-3mm}
As evident from Proposition~\ref{prop_1}, the computational efficiency of the  Bernoulli race clearly relies on the normalization constant $ Z(.)$. Note that the value of $ Z(.) $ could be interpreted as the average acceptance rate of PRC which depends on the hyper-parameter $ M(i,t-1)$. If the average acceptance rate for PRC for all particles is $\gamma$,  then we can express the expected number of iterations as $ \mathbb{E}[\Lambda^{i}_{t}] = \gamma^{-1} $. Therefore, the computational efficiency of Bernoulli race is similar to the PRC step and depends crucially on the hyper-parameter $ M $.

% ----------
\begin{thm}
For all $ K $, $ \exp (\hat{\mathcal{L}}_{\text{VRPF}}) $ is  unbiased for $ p_{\theta}(x_{1:T})$.
% , i.e.,  $ \mathbb{E} \left[\exp(\hat{\mathcal{L}}_{\text{VRPF}}) \right] =  p_{\theta}(x_{1:T})$.
Further, $ \mathbb{E}[\hat{\mathcal{L}}_{\text{VRPF}} ]  $ is non-decreasing in $ K $.

% satisfies the following properties.
% \begin{itemize}
%     \item $ \exp(\hat{\mathcal{L}}_{\text{VRPF}})$ is an unbiased estimator of $ p_{\theta}(x_{1:T}) $.
  
%      \item For finite K, $ \mathbb{E}[\hat{\mathcal{L}}_{\text{VRPF}};K]  \leq \log p_{\theta}(x_{1:T})   $
%     %\item As $ K \to \infty $ in Eq.~\eqref{iiw_mc_t}, $\mathbb{E}[ \hat{\mathcal{L}}_{\text{VRPF}}] \to \mathcal{L}_{\text{VRPF}} $
%     \item $ \mathbb{E}[\hat{\mathcal{L}}_{\text{VRPF}} ]  $ is non-decreasing in K.
% \end{itemize}
\label{prop_2}
\end{thm}
% ----------
%\begin{thm}
%For all $ K $, the Monte-Carlo estimator $ \hat{\mathcal{L}}_{\text{VRPF}} $ satisfies the following properties.
%\begin{itemize}
%    \item $ \exp(\hat{\mathcal{L}}_{\text{VRPF}})$ is an unbiased estimator of $ p_{\theta}(x_{1:T}) $.
  
%     \item For finite K, $ \mathbb{E}[\hat{\mathcal{L}}_{\text{VRPF}};K]  \leq \log p_{\theta}(x_{1:T})   $
    %\item As $ K \to \infty $ in Eq.~\eqref{iiw_mc_t}, $\mathbb{E}[ \hat{\mathcal{L}}_{\text{VRPF}}] \to \mathcal{L}_{\text{VRPF}} $
%    \item $ \mathbb{E}[\hat{\mathcal{L}}_{\text{VRPF}} ]  $ is non-decreasing in K.
%\end{itemize}
%\label{prop_2}
%\end{thm}
\vspace{-1mm}
The use of Monte-Carlo estimator in place of the true value of $ Z(.) $ creates an inefficiency, as depicted by Proposition~\ref{prop_2}. The monotonic increase in bound value with $ K $ is intuitive as we are constructing a more efficient estimator of $ Z(.) $ therefore getting a tighter bound. 
It is important to note that Algorithm~\ref{Algorithm_1} is producing an unbiased estimator of the marginal likelihood for all values of $ K $.

\begin{thm} \label{prop_3}
%Consider time $t-1$ at which we have uniformly weighted samples $ \{N^{-1},z^{ A^{i}_{t-1} }_{1:t-1}\}_{i=1}^{N}$ approximating $p_{\theta}(z_{1:t-1}|x_{1:t-1}) $.
Let the sampling distribution of the $ i^{th} $ particle (generated by PRC) at time $ t $ be $ r_{\theta,\phi}(z_{t}|z^{ A^{i}_{t-1} }_{1:t-1},x_{1:t})$, then
\vspace{-1.5mm}
\begin{eqnarray}
    KL\left( r_{\theta,\phi}(z_{t}|z^{ A^{i}_{t-1} }_{1:t-1},x_{1:t}) \,\|\,  p_{\theta}(z_{t}|z^{ A^{i}_{t-1} }_{1:t-1},x_{1:t}) \right) \\ \nonumber
    \leq KL\left( q_{\phi}(z_{t}|z^{ A^{i}_{t-1} }_{1:t-1},x_{1:t}) \,\|\,  p_{\theta}(z_{t}|z^{ A^{i}_{t-1} }_{1:t-1},x_{1:t}) \right).
\end{eqnarray}
\vspace{-5mm}
% Further,
% \begin{enumerate}
%     \item As $ M(i,t-1) $ $\to$ $0$,  $ r (z_{t}|z^{ A^{i}_{t-1} }_{1:t-1},x_{1:t}) $ $\to$ $ q(z_{t}|z^{ A^{i}_{t-1} }_{1:t-1},x_{1:t}) $.
%     \item As $ M(i,t-1) $ $\to$ $\infty $,  $  r(z_{t}|z^{ A^{i}_{t-1} }_{1:t-1},x_{1:t}) $ $\to$ $ p(z_{t}|z^{ A^{i}_{t-1} }_{1:t-1},x_{1:t}) $.
% \end{enumerate}
\end{thm}

Proposition~\ref{prop_3} implies that the use of the accept-reject mechanism within SMC refines the sampling distribution. Instead of accepting all samples, the PRC step ensures that only high-quality samples are accepted, leading to a tighter bound for VRPF in general (not always). We show in the supplementary material that  when $ M(i,t-1) $ $\to$ $\infty$, the PRC step reduces to pure rejection sampling~\cite{robert2013Monte}. On the other hand, $ M(i,t-1) $ $\to$ $ 0 $ implies that all samples are accepted from the proposal. Recall, $ M(i,t-1) $ is a hyperparameter that can be tuned to control the acceptance rate. For more details on tuning $ M $, see Section~\ref{tune_M}. %Both resampling and PRC are greedy processes; therefore, in some cases, the learned bound may be inferior to IWAE~\cite{burda2015importance}. However, experimental results indicate that our methodology is effective in general.

%Write that in prop3 shows that PRC always learns an improved postrior distr. as compared to standard pf. Write something that it will make the bound tigher in general. The role of M and how to tune it. 

 %First prove that marginal likelihood estimator is unbiased, prove proposition 1 of neurips submission (with resampling), Prove proposition 2 for full rejection control and describe how this property transfers sometimes to PRC framework (with resampling) as well. Overall there are three propositions now in the paper. Disuss the role of $N$, $ K $, and $ \gamma $.

 %We have to discuss why Bernoulli factory works, how it is able to generate unbiased ancestor variables.

\subsection{Gradient Estimation}\label{grad_sec}
%\textbf{extend this section}

For tuning the variational parameters, we use stochastic optimization. Algorithm~\ref{Algorithm_1} produces the marginal likelihood estimator by sequentially sampling the particles, ancestor variables, and particles for the normalization constant $(z^{1:N}_{1:T},A^{1:N}_{1:T-1} ,\delta^{1:N,1:K}_{1:T})$.

 \begin{figure*}
    \centering
    \includegraphics[width=14cm]{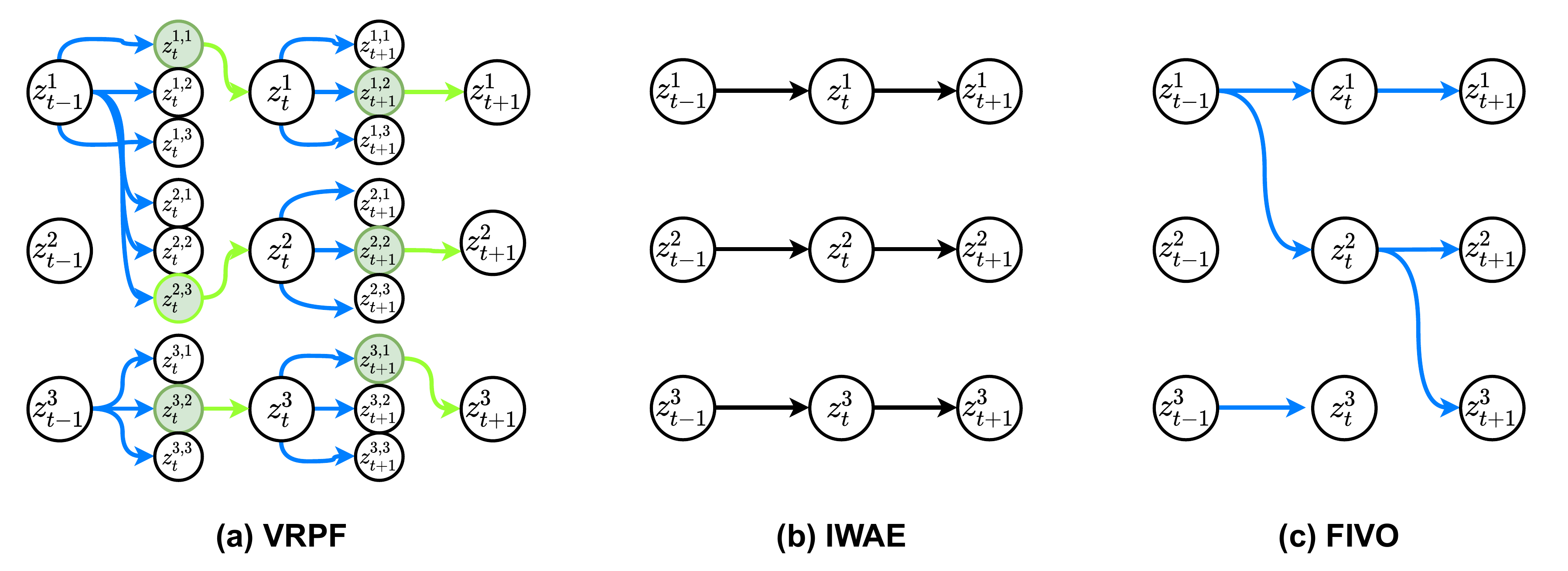}
    \caption{Comparison of VRPF with IWAE~\cite{burda2015importance} and FIVO~\cite{maddison2017filtering} (a) The blue arrows represent the resampling step, We then generate multiple samples from parametrized proposal $z^{i}_{t}|z^{i}_{1:t-1} $ out of which one sample is accepted via PRC, depicted via green arrows. (b) In IWAE, there is no resampling step and no PRC step (c). In FIVO, there is a resampling step (blue arrows) but no PRC step. } 
    \label{fig:VRPF_visual}
    \vspace{-2mm}
\end{figure*}

When the variational distribution $ q_{\phi}(.) $ is reparameterizable, we can make the sampling of $ \delta^{i,k}_{t} $ independent of the model and variational parameters. However, the generated particles $ z^{i}_{t} $ are not reparametrizable due to the PRC step. Finally, the ancestor variables are discrete and, therefore, cannot be reparameterized. The complete gradient can be divided into three core components (assuming  $ q_{\phi}(.) $ is reparametrizable) %\textbf{cite paper regarding reparametrization gradient explaining g-rep}: 
% \vspace{-1mm}
%
%\begin{equation}
%     \nabla_{\theta,\phi} \mathbb{E}[ \hat{\mathcal{L}}_{\text{VRPF}}]   =   g_{\text{rep}} + g_{\text{PRC}} + g_{\text{RSAMP}}, \label{biased_grad}
%\end{equation}
%where $ g_{\text{rep}}, g_{\text{PRC}},$ and $ g_{\text{RSAMP}}$ are as follows
\begin{eqnarray} 
& & \nabla_{\theta,\phi} \mathbb{E}[ \hat{\mathcal{L}}_{\text{VRPF}}]   =   g_{\text{rep}} + g_{\text{PRC}} + g_{\text{RSAMP}}, \label{biased_grad} \\ \nonumber
   & & g_{\text{rep}}  = \mathbb{E}_{ Q_{\text{VRPF} } } \left[ \nabla_{\theta,\phi}\hat{\mathcal{L}}_{\text{VRPF}} \right], \\ \nonumber
  & &   g_{\text{PRC}}  =  \mathbb{E}_{ Q_{\text{VRPF} } } \left[ \hat{\mathcal{L}}_{\text{VRPF}} \nabla_{\theta,\phi} \sum_{i=1}^{N}\sum_{t=1}^{T}\log r_{\theta,\phi}(z^{i}_{t}|.) \right], \\ \nonumber
&  &     g_{\text{RSAMP}}  =  \mathbb{E}_{ Q_{\text{VRPF} } } \left[ \hat{\mathcal{L}}_{\text{VRPF}} \nabla_{\theta,\phi} \sum_{i=1}^{N}\sum_{t=1}^{T-1} A^{i}_{t}\log \alpha_{t}    \right],% & \approx & g_{\text{rep}}.
\end{eqnarray}

%\vspace{-3mm}
where $ r_{\theta,\phi}(z^{i}_{t}|.) $ denotes the sampling density of the PRC step. Due to high gradient variance and intractability, we ignore $ g_{\text{PRC}}$ and $ g_{\text{RSAMP}}$ from the optimization. We have derived the full gradient and explored the gradient variance issues in the supplementary material. Please see Figure~\ref{GSSM_fig} (\emph{left}) comparing the convergence of biased gradient vs. unbiased gradients on a toy Gaussian SSM.  %For more details, please refer to the supplementary material.

\subsection{Learning the \emph{$\textbf{M}$} Matrix}\label{tune_M}
We use $ M $ as a hyperparameter for the PRC step which controls the acceptance rate of the sampler. The basic scheme of tuning $ M $ is as follows:
\vspace{-2mm}
\begin{itemize}
    %\item Set $ A^{i}_{t}=i $ if $ t \not \in \Lambda $ else $  A^{i}_{t}= A^{i}_{t} $
    \item Define a new random variable $F \big(z_{t+1}|z_{1:t}^{A^{i}_{t}},x_{1:t+1} \big)=$ 
    \[ \log \left(\frac{q_{\phi} \left(z_{t+1}|x_{1:t+1},z^{A^{i}_{t}}_{1:t} \right)}{p_{\theta} \left(x_{t+1},z_{t+1}|x_{1:t},z^{A^{i}_{t}}_{1:t} \right)} \right) .
    \]
    \item For $ j=1,2,\hdots,J $, draw \[z^{j}_{t+1} \sim q_{\phi} \left(z_{t+1}|x_{1:t+1},z^{A^{i}_{t}}_{1:t} \right) \,.
    \] 
    \item Evaluate $ \gamma \in [0,1] $ quantile value of $ \{F(z^{j}_{t+1}|.)\}_{j=1}^{J}$. In general for this case the acceptance rate would be around $ \gamma $ for all particles:
    \begin{equation}
        \log M(i,t) = -\mathcal{Q}_{F(z_{t+1}|z_{1:t}^{A^{i}_{t}},x_{1:t+1})}(\gamma). \label{M_val}
    \end{equation}
    
    \item If $ M $ matrix is very large then use a common $ \{M(.,t)\}_{t=1}^{T} $ for every time-step. In general, for this configuration, the acceptance rate would be \emph{greater} than equal to $ \gamma $ for all particles:
    \vspace{-1mm}
\begin{equation}
    \log M(.,t) = \min_{i=1, \dots, N} \left\{ -\mathcal{Q}_{F(z_{t+1}|z_{1:t}^{A^{i}_{t}},x_{1:t+1})}(\gamma) \right\}. \label{M_val_t}
\end{equation}
\end{itemize}
Through $\gamma$: a user parameter, we can directly control the acceptance rate. Therefore, both Bernoulli race and PRC would take around (less than) $\gamma^{-1} $ iterations to produce a sample for $ M $ value learned from \eqref{M_val} (see \eqref{M_val_t}). For implementation details please refer to the experiments. Algorithm~\ref{Algorithm_2} summarizes the complete optimization routine for the VRPF bound. We construct a stochastic gradient through \eqref{biased_grad} via a single draw from Algorithm~\ref{Algorithm_1}. To save time, we will update $ M $ once every $ F $ epochs. %The critical user parameters for setting up the optimization routine are $ K, \gamma,$ and $ F $. Please refer to the experiments section to learn more about setting these parameters.

%The complete optimization routine could be seen in Algorithm~\eqref{Algorithm_2}.\textbf{say few words about algorithm 2}

\begin{algorithm}[ht]
\caption{Optimization of VRPF lower bound }
\label{Algorithm_2}
\begin{algorithmic}[1]
\STATE{\textbf{Required}: $ p_{\theta}(x_{1:T},z_{1:T}) $, $ q_{\phi}(z_{1:T}|x_{1:T}) $, $ K$, $ F $, $\gamma $ }
\STATE{\textbf{Initialization}: $ M(i,t) = 0 \quad \forall i,t $  , $ ep = 0$ }
\WHILE{not converged }
\STATE{Compute $g_{\text{rep}}$ via Eq.~\eqref{biased_grad} }
\STATE{$(\theta_{t+1},\phi_{t+1})=(\theta_{t},\phi_{t})+\eta_{t} g_{\text{rep}}(\theta_{t},\phi_{t})$ }
\STATE{$ ep = ep +1 $}
\IF{ $ ep \hspace{1mm}\text{mod}\hspace{1mm} F ==0 $ }
\STATE{ Draw $ \{z^{j}_{t+1}\}_{j=1}^{J} \sim q_{\phi}(z_{t+1}|x_{1:t+1},z^{A^{i}_{t}}_{1:t})\hspace{1mm} \forall i,t $ }
\STATE{ Update $ M(i,t) $ from Eq.~\eqref{M_val} }
\ENDIF
\ENDWHILE
\STATE{\textbf{return} $(\theta^{*},\phi^{*})$ }
\end{algorithmic}
\end{algorithm}
\vspace{-2mm}

%Note that a similar scheme was also employed in~\cite{grover2018variational}. We update $ \{\{M(i,t-1)\}_{i=1}^{N}\}_{t=1}^{T} $ dynamically once every $F$ epochs to save time. To learn more on setting hyper-parameter $ M $, see~\cite{liu1998rejection,peters2012sequential}. %For certain cases the $ M $ matrix could be pretty large, therefore we will use a common $ \{M(.,t)\}_{t=1}^{T} $ for every time-step. In the Variational RNN experiments, we have chosen a single value for every time-step. Note that for this scheme we can ensure that the acceptance rate is always greater than equal to $ \gamma $ for all particles.  

%\vspace{-1mm}

 \section{Related Work and Special Cases}
 %\textbf{Can we discuss normalizing flows also??}
 
 There is a significant recent interest in developing more expressive variational posteriors for LVM. One way to address this is by employing richer variational families i.e. using hierarchical models~\citep{ranganath2016hierarchical}, copulas~\citep{tran2015copula}, or a sequence of invertible/non-invertible transformations~\citep{rezende2015variational,kumar2020regularized}. Another alternative, which is gaining attention lately is to combine VI with sampling methods. In particular, ~\citet{salimans2015markov,domke2017divergence,hoffman2017learning,li2017approximate,titsias2017learning,habib2018auxiliary,zhang2018ergodic,ruiz2019contrastive} use MCMC to construct a flexible VI bound. Other works uses \emph{approximate} rejection sampling in a variational framework~\cite{grover2018variational,gummadi2014resampled}. Apart from flexible bounds, approximations of sampling-based methods have also been employed to improve the generated images through GAN~\cite{goodfellow2014generative,azadi2018discriminator,turner2019metropolis,neklyudov2018metropolis}, construct richer priors for VAE~\citep{bauer2018resampled}, and improve gradient variance for VI~\cite{naesseth2016reparameterization}. %Recently a combination of VI with MCMC have also been employed to minimize $ KL(p||q) $ divergence~\citep{naesseth2020markovian} which unlike VI doesn't underestimate the posterior uncertainty. Recently there has been attempts to leverage particle smoothing within a variational framework also~\citep{lawson2018twisted}
 
 The other direction is to construct a particle approximation within VI. Specifically, IWAE~\cite{burda2015importance,domke2018importance} uses importance sampling (IS) to construct tighter VI bounds. Although IS is useful in many scenarios, SMC can yield significantly better estimates for sequential settings~\citep{doucet2009tutorial}. In particular~\citet{maddison2017filtering,naesseth2017variational,le2017auto,lawson2018twisted} have shown that SMC yield superior results than IS within a variational framework for sequential data. For a detailed review on particle approximations in VI please refer to~\citet{domke2019divide}.

 %There are two basic schemes for constructing tighter bounds on the log marginal likelihood:  sampling-based methods (MCMC, rejection sampling)~\cite{salimans2015markov,ruiz2019contrastive,hoffman2017learning,grover2018variational} or  multiple samples from VI distributions to increase the flexibility (IS, SMC)~\cite{burda2015importance,maddison2017filtering,lawson2018twisted,naesseth2015nested}.
 In this work, we present a unified framework for combining these two approaches, utilizing the best of both worlds. Although applying sampling-based methods on VI is useful, the density ratio between the true posterior and the improved density is often intractable. Therefore, we cannot take advantage of variance-reducing schemes like resampling, which is crucial for sequential models. We solve this issue through the Bernoulli race.
 
  \begin{figure*}[t]
	\begin{minipage}{0.4\linewidth}
		\centering
    \setlength{\tabcolsep}{1pt}
    \vspace{4mm}
        \begin{tabular}[b]{c c c c c}\toprule
         & \tiny{$ \mathbf{\log p_{\theta}(x_{1:T})} $ } & \tiny{\textbf{VSMC}}  & \tiny{$\mathbf{\gamma}=0.8$} & \tiny{$\mathbf{\gamma}=0.4$} \\ \midrule
      \small{\textbf{Case 1}} & \small{-18.27}  & \small{-25.78}  & \small{-24.80}  & \small{-21.91} \\ 
      \small{\textbf{Case 2}} &  \small{-84.33}  & \small{-230.46}  & \small{-197.15}   & \small{-187.25} \\
      \small{\textbf{Case 3}} &  \small{-33.89}  & \small{-159.96}   & \small{-108.47}  & \small{ -86.36} \\
      \small{\textbf{Case 4}} &  \small{-443.73}  & \small{-538.33}  & \small{-531.89}   & \small{-515.10} \\ \bottomrule
    \end{tabular}
    \includegraphics[width=6.0cm]{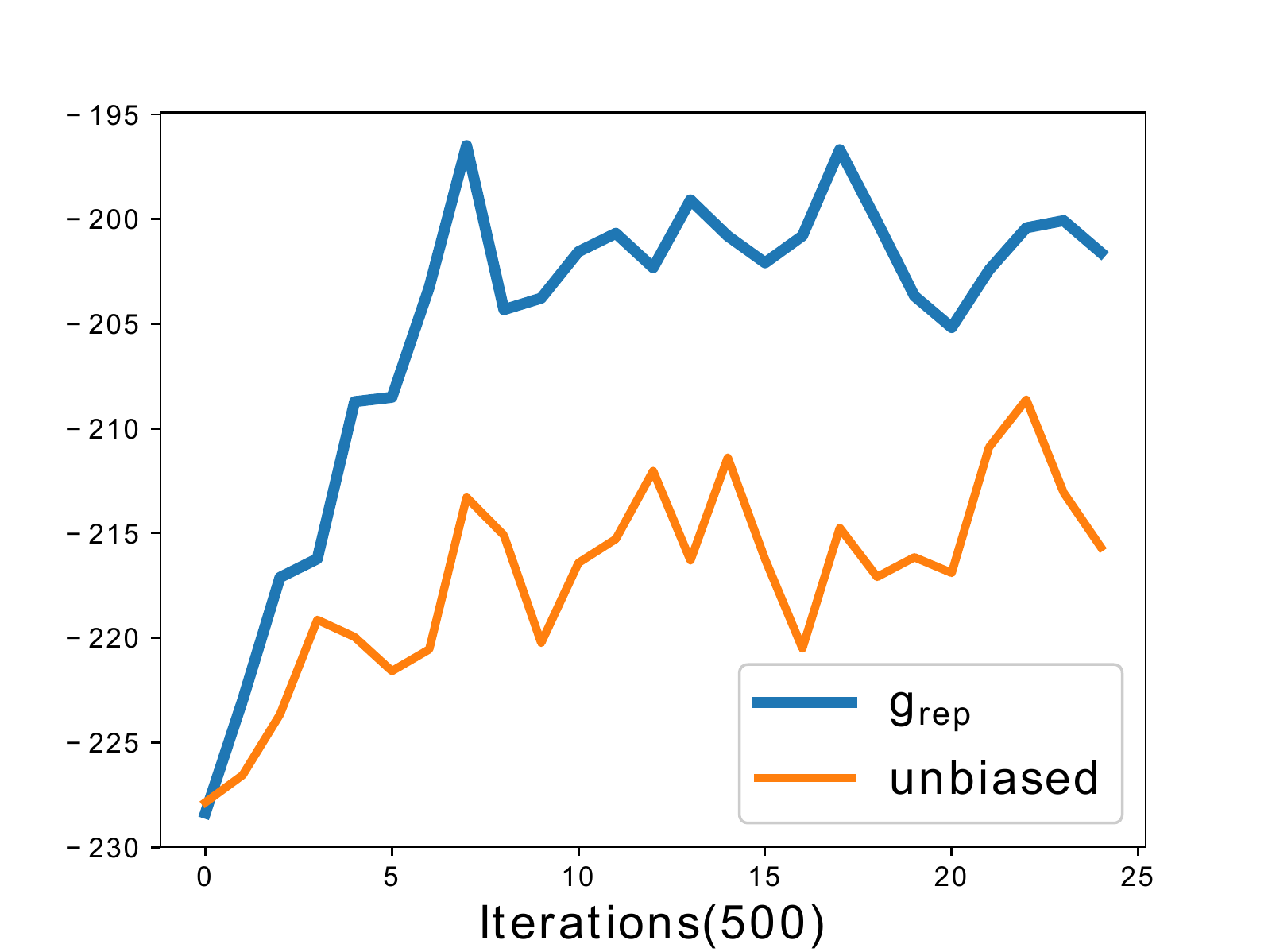}
	\end{minipage}%
	\begin{minipage}{0.6\linewidth}
		\begin{tabular}{c c}
		    \includegraphics[width=.47\linewidth]{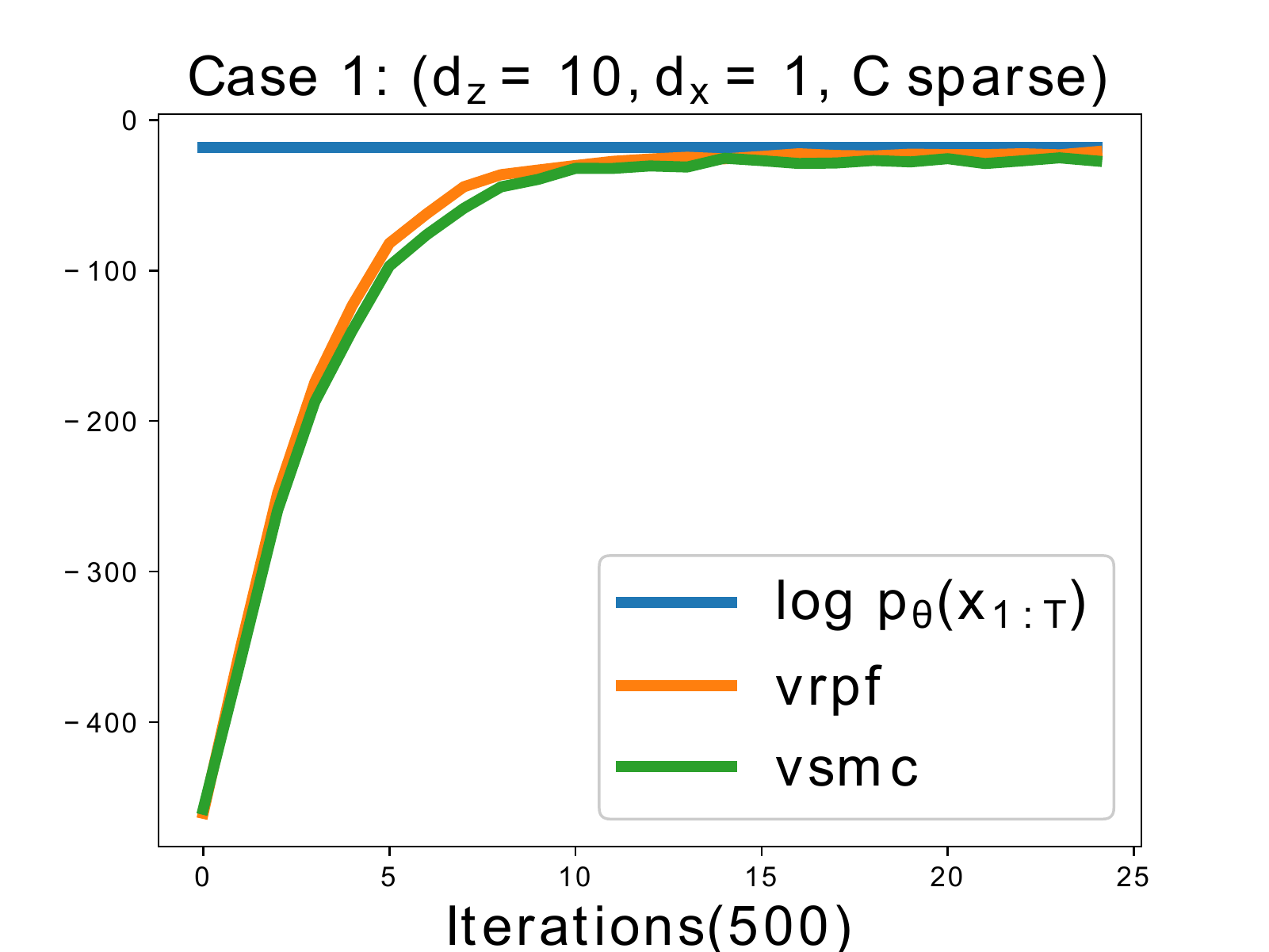} & \includegraphics[width=.47\linewidth]{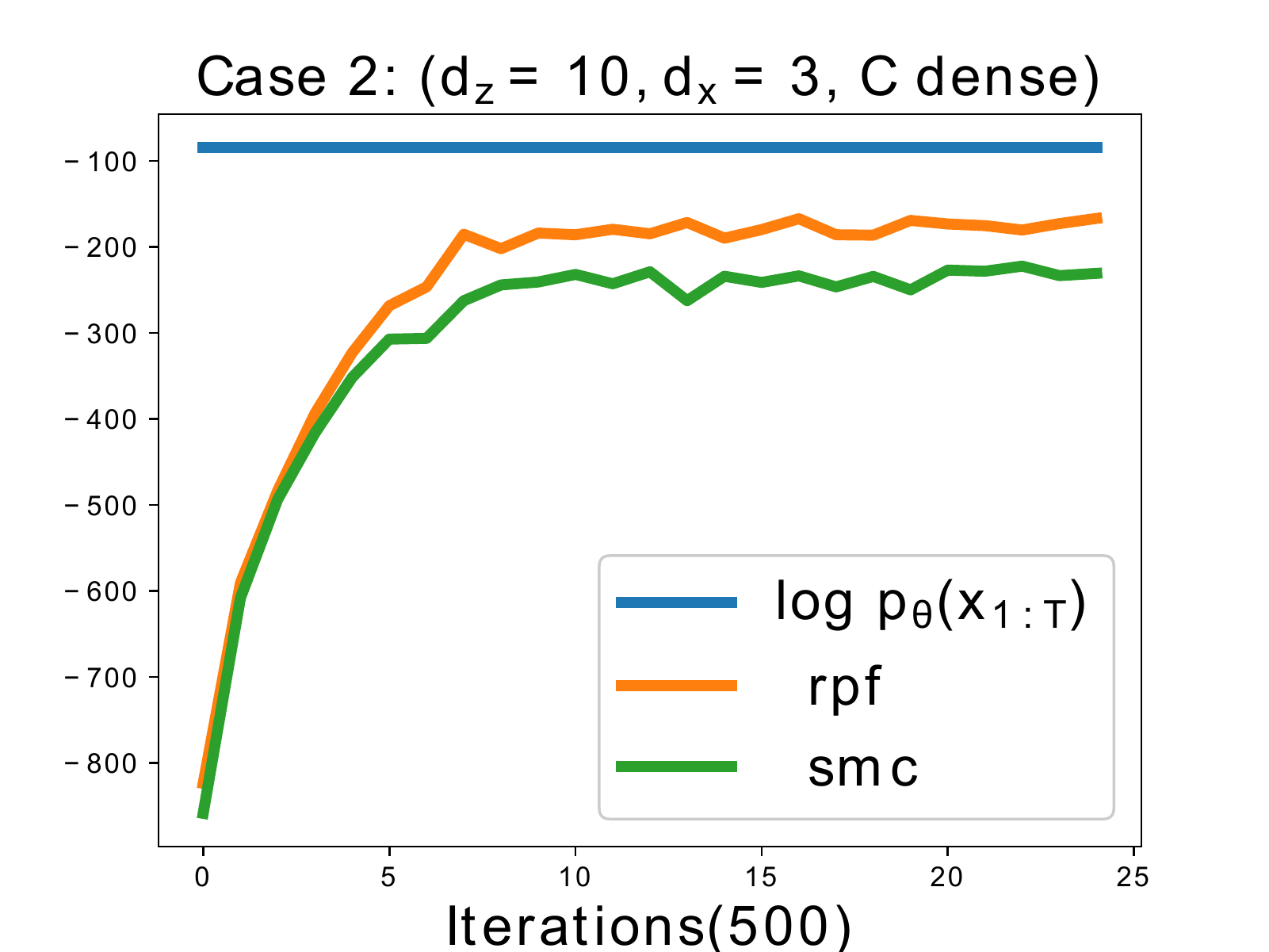}  \\
		    \includegraphics[width=.47\linewidth]{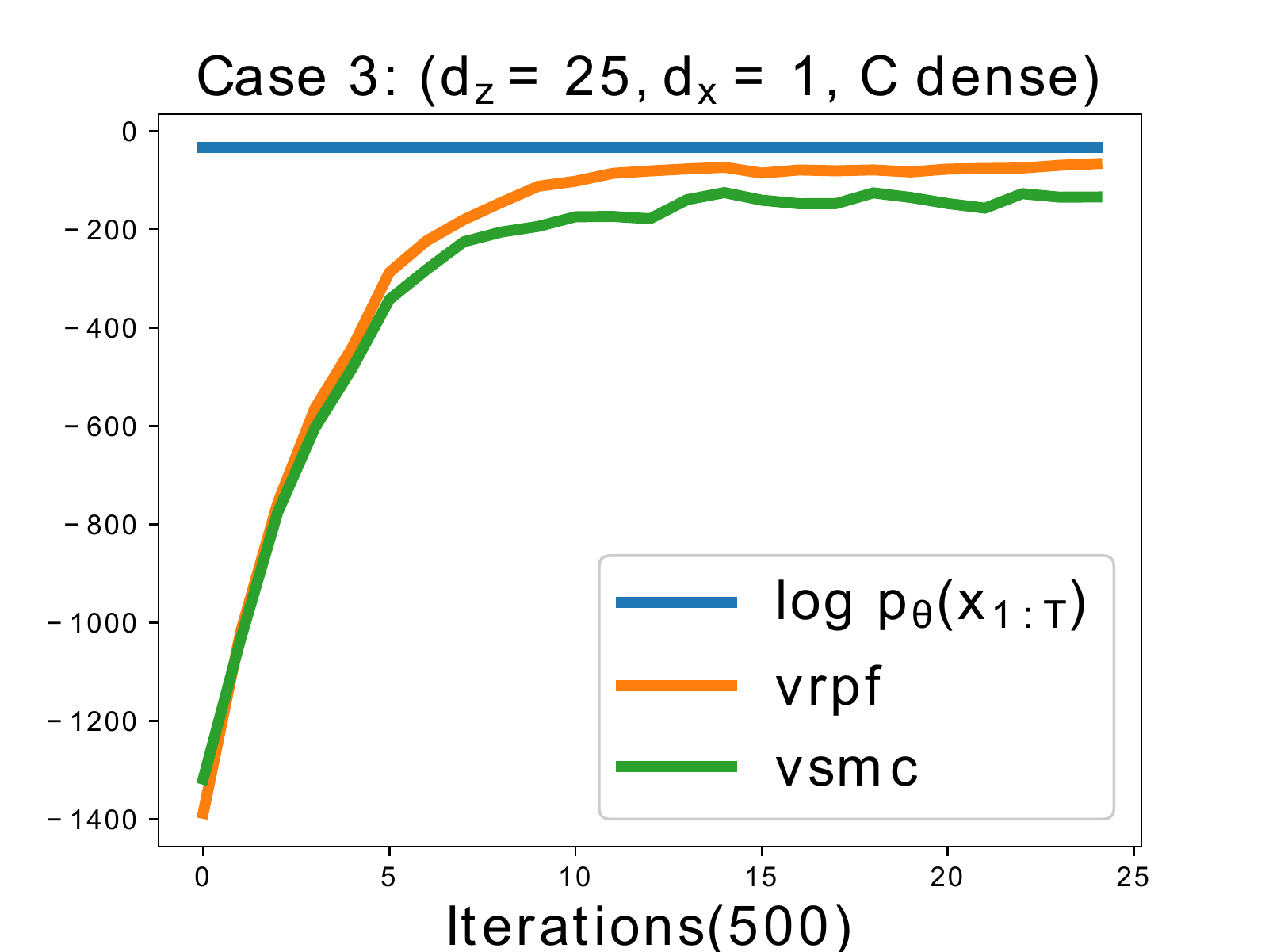}  & \includegraphics[width=.47\linewidth]{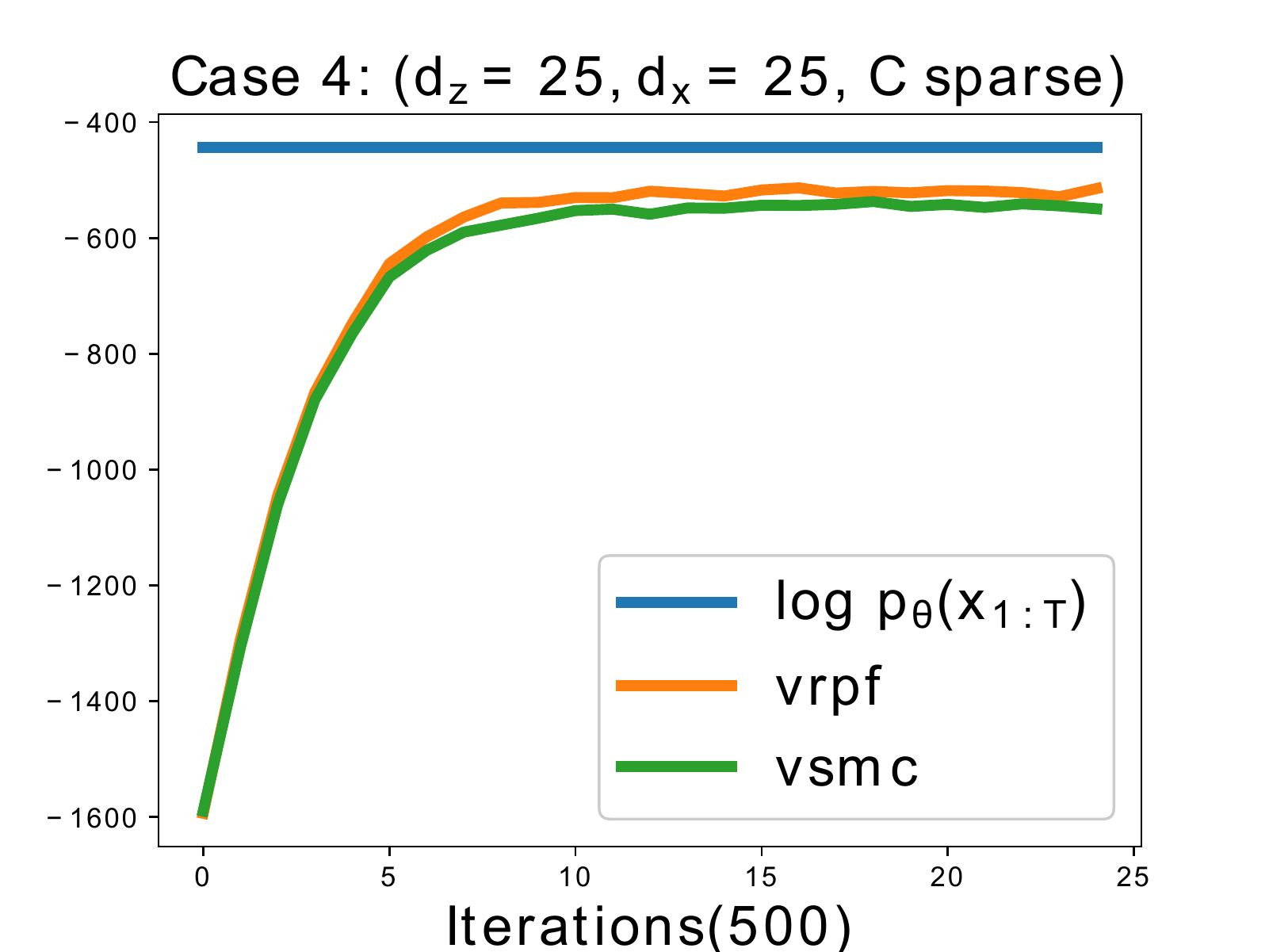} 
		\end{tabular}
	\end{minipage}
	\caption{(\emph{Left}) The figures compares the bound value for VRPF with full gradient and biased gradient~(\eqref{biased_grad}) as a function of iterations. (\emph{Left}) The Table compares the bound value for VSMC~\cite{naesseth2017variational} and VRPF for $ 80\%$ and $40\%$ acceptance rate. (\emph{Right}) We compare VSMC, VRPF ($40\%$ acceptance rate), and $ \log p_{\theta}(x_{1:T}) $ as a function of iterations. }
	\label{GSSM_fig}
\end{figure*}

 %\textbf{Discuss more VI+sampling work instead of Bernoulli factory and a slight discussion on Kudlicka and Schmon. Have a discussion more on VRS, VI+accept reject papers, VI+SMC papers and your contribution as a unification of the two}
 
Some closely related works to our method are VRS~\citep{grover2018variational} and FIVO~\citep{maddison2017filtering}. However, there are some key differences. In particular, consider a latent variable $\mathbf{z}_{1:t}=(z_{1},z_{2},\hdots,z_{t})$. In VRS, if the sample $\mathbf{z}^{i}_{1:t}$ is rejected, then we have to generate the \emph{entire} sequence of intermediate $z_{j}$'s again, which could be costly for a large probabilistic system with long sequences. However, if a sample is rejected for our method, we generate a new sample from a parametrized proposal $z_{t}|z^{i}_{1:t-1}$; therefore, we only introduce a \emph{partial} accept-reject at a local level, saving time. FIVO exploits the use of resampling within marginal likelihood estimation, thereby constructing a tight VI bound. However, in contrast to our method, it doesn't exploit sampling-based methods like rejection sampling; therefore, VRPF tends to form a tighter VI bound than FIVO. We found that even introducing a \emph{partial} accept-reject step with a high acceptance rate (more than 90 $\%$) is still useful. Please refer to Section~\ref{sec:vrnn} for more details.

%Secondly, we take advantage of the SMC approach instead, unlike VRS which uses IS and is not suitable for sequential models.\textbf{improve this paragraph}

A closely related work from SMC literature is BRPF~\citep{schmon2019bernoulli}, which also utilizes Bernoulli factories to implement unbiased resampling. BRPF provides a general unbiased SMC construction when the true weights are intractable. However, unbiasedness though important is still not sufficient for formulating a VI bound; we also need efficient implementation along with improved performance. Unlike the existing BRPF frameworks which are limited to niche one-dimensional toy examples, the specific framework of VRPF is important for VI due to several reasons. First, it unifies sampling-based method with a particle approximation giving us a flexible family of VI bounds. Second, it belongs to the general family of BRPF; therefore one can use Bernoulli race to perform unbiased resampling. Finally, Section~\ref{tune_M} demonstrates the efficient tuning of VRPF, thereby allowing us to scale our approach to general machine learning models like VRNN~\citep{chung2015recurrent} in contrast to the current BRPF frameworks. Another relevant work for unbiased estimation of SMC with PRC is that of~\citet{kudlicka2020particle}. In contrast to BRPF, this method samples one additional particle and keeps track of the number of steps required by PRC for every time-step to obtain their unbiased estimator. The weights are tractable for~\citet{kudlicka2020particle} as they do not take into account the effect of the normalization constant $ Z(.) $. It is important to note however that~\citet{kudlicka2020particle} does not consider the exact setting of~\citet{peters2012sequential}. Therefore, one cannot use~\citet{kudlicka2020particle} directly for VRPF in contrast to BRPF.  %Therefore, we are using BRPF for this paper.

 To provide more clarity, we will consider some special cases of VRPF bound and relate it with existing work: Note that for $ N=1 $ our method reduces to a special case of~\citet{gummadi2014resampled} which uses a constraint function $ C_{t}$ for every time-step and restarts the particle trajectory from $ \Delta_{t} $ (if $ C_{t}$ is violated). Therefore, if we use the setting $ C_{t}(z_{1:t}) = a(z_{t}|z_{1:t-1},x_{1:t} )$ and $ \Delta_{t} = t-1 $, we recover a particular case of~\citet{gummadi2014resampled}. For the special case of $ N=1$ and $ T=1$, our method reduces to VRS~\cite{grover2018variational}. For $ N, T>1$, if we remove the PRC step, our bound reduces to FIVO~\cite{maddison2017filtering}. Finally, if we remove both the PRC step and resampling, then our method effectively reduces to IWAE~\cite{burda2015importance}. Please refer to Figure~\ref{fig:VRPF_visual} for more details.

%The idea of combining variational inference with an inbuilt accept-reject mechanism is not new. In particular, the combination of MCMC, rejection sampling, and variational methods have been used in recent work to learn expressive variational posteriors \cite{salimans2015markov,ruiz2019contrastive,hoffman2017learning,titsias2017learning,li2017approximate,zhang2018ergodic,grover2018variational}. One closely related work to our method is Variational Rejection Sampling (VRS)~\cite{grover2018variational}. However, there are some key differences. In particular, consider a latent variable $\mathbf{z}_{1:t}=(z_{1},z_{2},\hdots,z_{t})$. In VRS~\cite{grover2018variational}, if the sample $\mathbf{z}^{i}_{1:t}$ is rejected then we have to generate the \emph{entire} sequence of intermediate $z_{j}$'s again which could be costly for a large probabilistic system with long sequences. On the other hand, if a sample is rejected for our method, we generate a new sample from a parametrized proposal $z_{t}|z^{i}_{1:t-1}$; therefore, we only introduce a \emph{partial} accept-reject at a local level, saving time. Secondly, we take advantage of the SMC~\cite{doucet2009tutorial} approach instead, unlike VRS~\cite{grover2018variational} which uses IS and is not suitable for sequential models.

\section{Experiments}\label{sec:exp}
We evaluate our proposed algorithm on synthetic as well as real-world datasets and compare them with relevant baselines. For the synthetic data experiment, we implement our method on a \emph{Gaussian SSM} and compare with VSMC~\cite{naesseth2017variational}. For the real data experiment, we train a \emph{VRNN}~\cite{chung2015recurrent} on the polyphonic music dataset.

\subsection{Gaussian State Space Model}
In this experiment, we study the linear Gaussian state space model. Consider the following setting %Assume that $ T=10 $ and number of particles as $ N =4 $, $z_{0}=0$ we have the following model.
\vspace{-1mm}
\begin{eqnarray*}
  z_{t}= Az_{t-1}+ e_{z}, \\
  x_{t} = Cz_{t} + e_{x},
\end{eqnarray*}
where $ e_{z}, e_{x} \sim \mathcal{N}(0,I) $ and $ z_{0} = 0$. We are interested in learning a good proposal for the above model. The latent variable is denoted by $z_{t}$ and the observed data by $x_{t}$. Let the dimension of $z_{t}$ be $d_{z}$ and dimension of $x_{t}$ be $d_{x} $. The matrix $A$ has the elements $(A)_{i,j}=\alpha^{|i-j|+1}$, for $\alpha = 0.42$.

We explore different settings of $ d_{z},d_{x}$, and matrix $ C $. A sparse version of $ C $ matrix measures the first $ d_{x} $ components of $ z_{t} $, on the other hand a dense version of $ C $ is normally distributed i.e $ C_{i,j} \sim \mathcal{N}(0,1) $. We consider four different configurations for the experiment. For more details please refer to Figure~\eqref{GSSM_fig}.

The variational distribution is a multivariate Gaussian with unknown mean vector $\mu=\{\mu_{d}\}_{d=1}^{d_{z}}$ and diagonal covariance matrix $\{\log\sigma_{d}^{2}\}_{d=1}^{d_{z}}$. We set $N=4$ and $T=10$ for all the cases:
\vspace{-1mm}
\begin{equation*}
    q(z_{t}|z_{t-1})\sim \mathcal{N}\left(z_{t}| A z_{t-1}+\mu,\text{diag}(\sigma^{2}) \right). \label{prop_ssm}
\end{equation*}

%\textbf{Case 1}:  Sparse version of $ C $, $d_{z}=10$, $d_{x}=1$, Acceptance rate $\gamma \in \{0.8,0.4\}$.

%\textbf{Case 2}: Dense version of $ C $ i.e $ C_{i,j} \sim \mathcal{N}(0,1) $, $d_{z}=10$, $d_{x}=3$, Acceptance rate $\gamma \in \{0.8,0.4\}$.

%\textbf{Case 3}:  Dense version of $ C $ i.e $ C_{i,j} \sim \mathcal{N}(0,1) $, $d_{z}=25$, $d_{x}=1$, Acceptance rate $\gamma \in \{0.8,0.4\}$.

%\textbf{Case 4}:  Sparse version of $ C $, $d_{z}=25$, $d_{x}=25$, Acceptance rate $\gamma \in \{0.8,0.4\}$. \{\{M(i,t-1)\}_{i=1}^{N}\}_{t=1}^{T}

The $ M $ matrix (see Eq.~\eqref{M_val}) for approximate rejection sampling is updated once every $ F=10 $ epochs with acceptance rate $\gamma \in \{0.8,0.4\}$. For estimating the intractable normalization constants, we set $ K=3 $. Figure~\ref{GSSM_fig}: (\emph{left}) compares the convergence of biased gradient vs unbiased gradients. Note that we obtain a much tighter bound as compared to VSMC~\cite{naesseth2017variational}.

% \textbf{Extend this section till next page}

 \subsection{Variational RNN}\label{sec:vrnn}
 VRNN~\cite{chung2015recurrent} comprises of three core components: the observation $x_{t}$, stochastic latent state $z_{t}$, and a deterministic hidden state $ h_{t}(z_{t-1},x_{t-1},h_{t-1}) $, which is modeled through a RNN. For the experiments, we use a single-layer LSTM for modeling the hidden state. %For a length $ T $ sequence, the variational distribution and joint data likelihood are defined as follows
%\begin{equation}
%   r(z_{1:T}|x_{1:T}) = \frac{\prod_{t=1}^{T} q_{t}(z_{t}|h_{t}(z_{t-1},x_{t-1},h_{t-1}),x_{t})a_{t}(z_{t}|h_{t}(z_{t-1},x_{t-1},h_{t-1}),x_{t}) }  {\prod_{t=1}^{T} Z_{t}(h_{t}(z_{t-1},x_{t-1},h_{t-1}),x_{t}) } \label{var_dist_VRNN}
%\end{equation}
%Note that $ a_{t}(z_{t}|.) $ is the acceptance probability for the PRC step~\eqref{accept_prob} and $ Z_{t}(.)$ is the intractable normalization constant. Similarly, we can write down the joint data likelihood as 
%\begin{equation}
%p(z_{1:T},x_{1:T})= \prod_{t=1}^{T} p_{t}(z_{t}|h_{t}(z_{t-1},x_{t-1},h_{t-1}),x_{t})g_{t}(x_{t}|h_{t}(z_{t-1},x_{t-1},h_{t-1}),z_{t}) \label{joint_lik_VRNN} 
%\end{equation}
The conditional distributions $p_{t}(z_{t}|.)$ and $q_{t}(z_{t}|.)$ are assumed to be factorized Gaussians, parametrized by a single layer neural net. The output distribution $g_{t}(x_{t}|.)$ depends on the dataset. For a fair comparison, we use the same model setting as employed in FIVO~\cite{maddison2017filtering}. We evaluate our model on four polyphonic music datasets: \texttt{Nottingham}, \texttt{JSB chorales}, \texttt{Musedata}, and \texttt{Piano-midi.de}~\citep{boulanger2012modeling}.

 \vspace{-2mm}
\begin{table*} [!htbp]
  \small
\centering
\caption{We report Test log-likelihood for models trained with FIVO, IWAE, ELBO, and VRPF. For VRPF $ N=(4,6) $ and  $ (K,\gamma) \in \{(1,0.9),(1,0.8),(3,0.9),(3,0.8)\}$ (results are in this order). The results for pianoroll data-sets are in nats per timestep.}
\label{Table_3}
\vskip 0.15in
\begin{tabular}{ c  c  c c c |c c c c c}
%\hline
%    \multicolumn{1}{c}{\small{N} }  & \multicolumn{1}{c}{\small{Data}} & \multicolumn{1}{c}{\small{ELBO}} & \multicolumn{1}{c}{ \small{IWAE}}  & \multicolumn{1}{c|}{ \small{FIVO}} & \multicolumn{4}{c}{ \small{VRPF}}   \\
%      & \small{Nott}  &  & \small{-3.10} & \small{-3.96} & \small{-3.00}  & \small{-3.09} & \small{-3.05} & \small{-3.11} \\ 
%  2 & \small{jsb}  &  & \small{-9.42}  & \small{-9.42} & \small{-7.74} & \small{-7.28}  & \small{-7.42} & \small{-7.28} \\ 
%        & \small{Piano}  &  &  & \small{-9.84} & \small{-10.38}  & & & \\
%         & \small{Muse}  &  & \small{-7.50} & \small{-7.42} & \small{-7.41}  & \small{-6.83} & & \\ 
\toprule
   \multicolumn{1}{c}{ \textbf{N}  }  & \multicolumn{1}{c}{ \textbf{Data} } & \multicolumn{1}{c}{\textbf{ELBO}} & \multicolumn{1}{c}{ \textbf{IWAE}  }  & \multicolumn{1}{c|}{ \textbf{FIVO} } & \multicolumn{1}{c}{ \textbf{N}  } & \multicolumn{4}{c}{ \textbf{VRPF}    }   \\
   \midrule
    &  Nott  & -3.87 & -3.12   & -3.07  &  & \textbf{-2.96}   & -2.98  &  -2.99  & -2.96  \\ 
  & jsb  &  -8.69 & -8.01   & -7.51   & & -7.41  & \textbf{-7.28}  & -7.37  & -7.36    \\ 
 5     &   Piano  & -7.99   &  -7.97  &  -7.85 & 4   &  -7.82  & -7.86  & \textbf{-7.80}  & -7.85  \\
       &   Muse  & -7.48   & -7.45 & -6.75  &  & -6.61   & -6.63  & -6.66  & \textbf{-6.58}  \\ 

%\multicolumn{2}{c}{ Avg. Rank  } &   &   &  & 2.75 & 2.5 & 2.5 & 1.5  \\ 
\toprule
    \multicolumn{1}{c}{ \textbf{N} }  & \multicolumn{1}{c}{ \textbf{Data}  } & \multicolumn{1}{c}{  \textbf{ELBO}} & \multicolumn{1}{c}{ \textbf{IWAE} }  & \multicolumn{1}{c|}{ \textbf{FIVO} } & \multicolumn{1}{c}{ \textbf{N}  } & \multicolumn{4}{c}{ \textbf{VRPF} }   \\
    \midrule
      & Nott  & -3.87   & -3.87  & -2.99 &  & -2.93   & -2.93  & \textbf{-2.90}  & -2.91 \\ 
       & jsb  & -8.69  & -8.32  & -7.40 & & -7.29   & -7.21  & -7.16  & \textbf{-7.14} \\ 
     8   & Piano  &  -7.99  & -8.04  & -7.80 & 6  & -7.78   & -7.77  & -7.79  & \textbf{-7.77} \\
         & Muse & -7.48   & -7.41  & -6.67 &  & -6.60  & \textbf{-6.57}  & -6.61  & -6.60 \\ 
         \toprule
\multicolumn{2}{c}{ \textbf{Avg. Rank}  } & \scriptsize{6.87 $\pm$ 0.33}   & \scriptsize{6.12 $\pm$ 0.33}   & \scriptsize{4.87 $\pm$ 0.33} &  & \scriptsize{2.87 $\pm$ 1.05} & \scriptsize{2.62 $\pm$ 1.21}  & \scriptsize{2.87 $\pm$ 1.26}  & \textbf{\scriptsize{1.75 $\pm$ 0.66} } \\ 
\bottomrule
\end{tabular} 
\vskip -0.1in
\end{table*}

Each observation $x_{t}$ is represented as a binary vector of 88 dimensions. Therefore, we model the observation distribution $g_{t}(x_{t}|.)$ by a set of 88 factorized Bernoulli variables. We split all four data-sets into the standard train, validation, and test sets. For tuning the learning rate, we use the validation test set. Let the dimension of hidden state (learned by single layer LSTM) be $d_{h}$ and dimension of latent variable be $d_{z}$. We choose the setting $d_{z}=d_{h}=64$ for all the data-sets except JSB. For modeling JSB, we use $d_{z}=d_{h}=32$. For VRPF we have considered $ N \in \{4,6\} $ Further, for every $ N $, we consider four settings $ (K,\gamma) \in \{(1,0.9),(1,0.8),(3,0.9),(3,0.8)\}$. The $ M $ hyper-parameter for the PRC step is learned from Eq.~\eqref{M_val_t} due to large size and is updated once every $F=50$ epochs. Note that in this scenario, the acceptance rate for all particles would be \emph{greater} than equal to $ \gamma $. For more details on experiments, please refer to the supplementary material.

As discussed in Section~\eqref{theore_prop}, the PRC step and Bernoulli race have time complexity $ \mathcal{O}(N/\gamma)$ for producing $ N $ samples (assuming average acceptance rate $\gamma$). Therefore, we consider $\lceil N\gamma^{-1}\rceil  $ particles for IWAE and FIVO to ensure effectively the same number of particles, where $ N \in \{4,6\}$ and $ \gamma  = 0.8 $. Note, however, that the acceptance rate is $\geq \gamma$, so this adjustment actually favors the other approaches more. For FIVO, we perform resampling when ESS falls below $N/2$. Table~\ref{Table_3} summarizes the results which show whether rejecting samples provide us with any benefit or not, and as the results show, our approach, even with the aforementioned adjustment, outperforms the other approaches in terms of test log-likelihoods.%, while still having a similar computational cost.

%In Section~\ref{theore_prop}, we discussed the effect of $ K $ and PRC rejection rate on VRPF bound. We expect a performance improvement when $ K $ and the rejection rate is increased. Although the results for VRPF's different configurations are almost the same, we still get the best average ranking for $ (K=3,\gamma=0.8) $. Overall, for most cases, VRPF bound performs better than FIVO~\cite{maddison2017filtering} and IWAE~\cite{burda2015importance} for a variety of configurations.

In VRPF, improvement in the bound value comes at the cost of estimating the normalization constant $ Z(.) $, i.e., $ K$. On further inspection, we can clearly see that increasing $ K $ doesn't provide us with any substantial benefits despite the increase in computational cost. Therefore, to maintain the computational trade-off $ (K=1,\gamma>0.8) $ seems to be a reasonable choice for VI practitioners. 

Table~\ref{Table_3} signifies that rejecting samples with low importance weight is better instead of keeping a large number of particles (at least for a reasonably high acceptance rate $\gamma$). It is interesting to note that \emph{partial} accept-reject indeed helps empirically. In the above VRNN experiment the latent variable $ \mathbf{z}_{1:T} $ is fairly high-dimensional for example: in \texttt{Piano-midi.de} the maximum sequence length is of order $ 10^{6}$. Therefore, it is straightforward to verify that one cannot use sampling-based methods directly. 

Our experimental results indicate that \emph{partial} sampling is useful even for high-dimensional applications. Specifically, we want to emphasize that accept-reject though useful, is still limited in its nature compared to MCMC algorithms like Hamiltonian Monte Carlo (HMC)~\cite{neal2011mcmc}. Note, however, we are still getting improved results for such large sequences despite using accept-reject. Therefore, exploring the general area of \emph{partial}/greedy sampling within high-dimensional time series models would make for interesting future work.

%The proposed bound uses more particles (PRC step and Bernoulli race) than existing approaches like FIVO and IWAE due to intractability.

\vspace{-2mm}
\section{Conclusion}
We introduced VRPF, a novel bound that combines SMC and \emph{partial} rejection sampling with VI in a synergistic manner. This results in a robust VI procedure for sequential latent variable models. Instead of using standard sampling algorithms, we have employed a partial sampling scheme suitable for high dimensional sequences. Our experimental results clearly demonstrate that VRPF outperforms existing bounds like IWAE~\cite{burda2015importance} and standard particle filter bounds~\cite{maddison2017filtering,naesseth2017variational,le2017auto}.  Future work aims at designing a scalable implementation for VRPF bound that consumes fewer particles and exploring \emph{partial} versions of powerful sampling algorithms like HMC instead of rejection sampling.

\nocite{langley00}

\bibliography{example_paper}
\bibliographystyle{icml2021}


\end{document}